\documentclass[journal]{IEEEtran}
\usepackage{blindtext}
\usepackage{graphicx}
\usepackage{makeidx}
\usepackage{float}
\usepackage{multirow}
\usepackage{subfig}
\usepackage{adjustbox}
\usepackage{amsmath}
\ifCLASSINFOpdf
\else
\fi
\begin{document}
%
\title{Semi-supervised Learning using Denoising Autoencoders for Brain Lesion Detection and Segmentation }
%
%
%

\author{
    \IEEEauthorblockN{Varghese Alex\IEEEauthorrefmark{1}, Kiran~Vaidhya\IEEEauthorrefmark{1}, Subramanium~Thirunavukkarasu\IEEEauthorrefmark{1}, Kesavdas Chandrasekharan\IEEEauthorrefmark{2},
 Ganapathy~Krishnmurthi\IEEEauthorrefmark{1}}\\
    \IEEEauthorblockA{\IEEEauthorrefmark{1}Indian Institute of Technology Madras, Chennai}
 \\ \IEEEauthorblockA{\IEEEauthorrefmark{2}Sree Chitra Tirunal Institute for Medical Sciences and Technology, Thiruvananthapuram}}
\maketitle

\begin{abstract}
The work presented explores the use of denoising autoencoders (DAE) for brain lesion detection, segmentation and false positive reduction. Stacked denoising autoencoders (SDAE) were pre-trained using a large number of unlabeled patient volumes and fine tuned with patches drawn from a limited number of patients (n=20, 40, 65). The results show negligible loss in performance even when SDAE was fine tuned using 20 patients. Low grade glioma (LGG) segmentation was achieved using a transfer learning approach wherein a network pre-trained with High Grade Glioma (HGG) data was fine tuned using LGG image patches. The weakly supervised SDAE (for HGG) and transfer learning based LGG network were also shown to generalize well and provide good segmentation on unseen BraTS 2013 \& BraTS 2015 test data. An unique contribution includes a single layer DAE, referred to as novelty detector(ND). ND was trained to accurately reconstruct non-lesion patches using a mean squared error loss function. The reconstruction error maps of test data were used to identify regions containing lesions. The error maps were shown to assign unique error distributions to various constituents of the glioma, enabling localization. The ND learns the non-lesion brain accurately as it was also shown to provide good segmentation performance on ischemic brain lesions in images from a different database. 
\end{abstract}


\IEEEpeerreviewmaketitle

\section{Introduction}

This paper Gliomas are a type of primary brain tumor that affect the glial cells in the brain. Based on severity, gliomas are further divided to HGG and LGG. Automatic segmentation of Gliomas from MRI, a preliminary step for treatment planning and determining disease progression, is a challenging task due to heterogeneity of tissue within the lesion, non uniform dynamic range of MR images and diffused borders of tumors. Furthermore multiple MRI sequences, namely T1, T2, FLAIR and T1 post contrast (T1c) are required for accurate segmentation. These sequences provide complementary information about the lesion. For e.g., T2 weighted sequence and FLAIR helps in segmenting the  gross tumor while T1 post contrast sequence helps to delineate the enhancing tumor and necrotic region from gross lesion. A fully automated image segmentation pipeline is thus necessary for evaluating large number of patients across multiple centers.\\
\subsection{Literature Survey}
 In the recent past, various fully automated techniques have been proposed to segment Gliomas and they can be broadly classified as either generative or discriminative techniques \cite{IEEEhowto:benchmark}. 
Generative techniques model the joint distribution of the voxel classes and voxel specific features. A typical approach is to register the images onto an probabilistic atlas \cite{IEEEhowto:gen2}-\cite{IEEEhowto:gen5}. An atlas represents a normal healthy brain and comprises of white matter, gray matter, ventricles, brain stem etc. Following registration various techniques have been developed to classify the tumor as an outlier/additional class. For e.g. \cite{IEEEhowto:gen5} used Covariance Determinant estimator to detect outliers followed by further segmentation using K-Means algorithm. Since the presence of large tumors or resectional cavities alter the structure of brain, the performance of generative models can be impacted by the registration technique used to align images and spatial priors \cite{IEEEhowto:gen6}. Overall, as stated in \cite{IEEEhowto:benchmark}, \cite{IEEEhowto:gen5}-\cite{IEEEhowto:gen7}, generative techniques perform well on unseen data. A recent work \cite {IEEEhowto:gen6} on a hybrid generative/discriminative model for glioma segmentation shows a boost in performance by combining a generative and discriminative approach. 

\par Discriminative techniques model/determine the class conditional distribution given the image features for eg. voxel intensities. Discriminative techniques such as Random Forest \cite{IEEEhowto:rf1}-\cite{IEEEhowto:rf8}  and  Support Vector Machines \cite {IEEEhowto:svm2}, \cite{IEEEhowto:svm1} have been applied to Brain Tumor segmentation. These techniques are suited for multiclass problems and  uses  hand coded features such as mean, median, skewness, symmetric of the brain to name a few, to classify voxels.  Discriminative techniques tend to misclassify certain voxels as lesion at anatomically and physiologically  unlikely locations since each voxel is modeled to be independent from its neighbouring voxels \cite{IEEEhowto:dl1}. However, Conditional Random Fields and Markov Random fields can be used to regularize the segmentation and could lead to improved results. The overall performance of discriminative techniques  in general would depend on the quality of the computed features. 
 
\par 
In the past decade, deep learning techniques such as Deep Belief networks, Convolutional Neural Networks, Stacked Denosing Autoencoders have been used in a variety of image classification and segmentation task \cite{IEEEhowto:dl2}-\cite{IEEEhowto:dl5}. Deep learning techniques are capable of learning features such as edges, textures, patterns and various higher order features from raw images. Recently convolutional neural networks (CNNs) have been used for segmentation of gliomas from MR images \cite{IEEEhowto:dl1}- \cite{IEEEhowto:dl11} and have outperformed other fully automatic techniques. CNNs can be considered as discriminative models since they predict the posterior probability given the image features. Typically, a large number of labeled data is required to train discriminative models especially deep learning based ones like CNNs. 
\par Among deep learning models, Restricted Boltzman Machines(RBM) and Deep Boltzman Machines \cite{IEEEhowto:dbm} can be considered as generative models. Convolutional RBMs \cite{IEEEhowto:crbm} have been used to extract features to aid semi-automated Glioma segmentation and was judged the best entry in the BraTS 2015 challenge. The focus of this paper is on SDAEs that learn a compact encoding of the data which can then be used as features for classification. Autoencoders and its variants which include SDAE's and sparse stacked autoencoders have been in various medical image processing applications \cite{IEEEhowto:sdae2}-\cite{IEEEhowto:ssae}. Both RBMs and autoencoders can be pre-trained using unlabeled data. RBMs being energy based models are trained using Markov chain techniques like Gibbs sampling, while autoencoders, DAE \& SDAE have the the advantage of being trained using gradient based backpropagation techniques. 

 \par One of the major issues that arise in training deep networks is class imbalance. Class imbalance is particularly acute in medical imaging problems since lesions constitute a miniscule percentage of image voxels. For instance in gliomas, lesion voxels form less than 2\% of the total number of image voxels, in such scenarios a novelty/anomaly detection approach would be very effective. The principles of anomaly detection are well studied and typically involves detecting outliers or rare events by measuring a distance metric obtained from a parametric model of the data (excluding the anomalies). Autoencoders  and other machine learning techniques \cite {IEEEhowto:novel}, \cite{IEEhowto:anamoly} have also found applications in novelty detection but have not been explored in the context of brain lesion detection from multi-sequence MR images. In the next section the paper's original contribution is outlined based on the unsupervised training and novelty detection approach for glioma segmentation and ischemic lesion segmentation. 

\subsection{Contribution}
This paper describes the application of denoising autoencoders for the detection and segmentation of brain lesions from multi-sequence MR images. Specifically the contributions are:
\begin {itemize}
\item False positive reduction for gliomas and candidate detection for brain lesions (Ischemic lesion) using a novelty detector.
\item Variant of the Novelty Detector, called Cascaded Novelty Detector (CND) which generates unique error distributions for various constituents of glioma.
\item Demonstrating Semi supervised and weakly supervised learning by training SDAE using patches drawn from limited patient volumes (n=20).
\item Transfer learning approach for LGG, which had limited number of labeled training data. The LGG network was obtained by fine tuning the  pre trained HGG network.

\end {itemize}

The manuscript is laid out as follows. Section II describes the data set used, section III describes the  preprocessing of image data, training of SDAE's and post processing using ND. Section IV describes the results and discusses the performance of SDAE's on test data for the brain tumor segmentation task. The paper concludes with the summary and discussion of future direction in  Section V.

\section{Training Data}
The publicly available  BraTS-2015 data set  \cite{IEEEhowto:benchmark}, \cite{IEEEhowto:brats2015} was used for training the networks. The data set comprises of 220 HGG and 54 LGG patient data. The HGG data set is composed of patients imaged only  once (single time point) and patients who are scanned multiple times (longitudinal data). The HGG data set comprises of 123 single time point patients (ST) and 97 longitudinal patients (LT), while no longitudinal studies were found in  the LGG data set. Each patient data comprises of a FLAIR, T2 weighted, T1 weighted and T1 post contrast sequence. Each voxel in the image volumes is classified as one of the five classes namely Normal, Edema, Non Enhancing Tumor, Necrotic Region and Enhancing Tumor as summarized in Table \ref{tabs:class}. There exists huge data imbalance among classes in both HGG and LGG data set (Table \ref{data imbalance}). Furthermore, certain classes occur more frequently in one grade of Glioma than the other, for e.g. enhancing tumor and necrotic region is more prominent in HGG while non enhancing tumor is more prominent in LGG.

\begin{table}[h]
\centering
\caption{Labels associated to various types of Lesion in the images}
\begin{tabular}{|c|c|}
\hline
Type of Lesion  & Class \\
\hline
Healthy/No Lesion and Background & 0 \\
\hline
Necrotic Region & 1 \\
\hline
Edema & 2 \\
\hline
Non Enhancing Tumor & 3 \\
\hline
Enhancing Tumor & 4\\
\hline
\end{tabular}
\label{tabs:class}
\end{table}

\begin{table}[h]
\centering
\caption{Amount of Data imbalance in \% in HGG and LGG }
\begin{tabular}{|c|c|c|c|c|c|}
\hline
Grade of Glioma  & class 0 & class 1& class 2 & class 3 &class 4 \\
\hline
HGG & 98.74 & 0.29 &0.59 & 0.0091 & 0.35 \\
\hline
LGG & 98.79 & 0.05 &0.8 & 0.34 & 0.0013 \\
\hline
\end{tabular}
\label{data imbalance}
\end{table}
\par The ischemic lesion training database made available as part of the ISLES 2015 challenge \cite{IEEEhowto:isles} was used for demonstrating candidate lesion detection using ND. The database consists 28 patient volumes comprising of Diffusion weighted images, FLAIR, T1 and T2 weighted sequences. 
\section{Methods}
\subsection {Background}

Autoencoders are neural networks that were originally used for dimensionality reduction. They are trained to reconstruct the input data and dimensionality reduction is achieved by using lesser number of neurons in the hidden layer than in the input layer. A deep autoencoder is obtained by stacking multiple layers of encoders with each layer trained independently (pre-training) using an unsupervised learning criterion. A classification layer can be added to the pre-trained encoder and further trained with labeled data (fine tuning). Such an approach initially outlined in \cite{IEEEhowto:dl4} was shown to be an effective way to train deep networks. Denoising autoencoder is a variant where the hidden layer is pre-trained with artificially corrupted data and the reconstruction error is calculated against the uncorrupted data. DAEs provide robust features which in turn improves the classification accuracy \cite{IEEEhowto:dl5}. 
\subsection {Overview}
\par  In this work a single layer denoising autoencoder was used as an anomaly/novelty detector by training the network to reconstruct non-lesion patches. The reconstruction error  corresponding to lesion and non lesion patches would then be significantly different. 
\par  SDAEs were pre-trained layer by layer using a large number of unlabeled patches. The network was fine tuned using labeled patches drawn from a limited subset of patients after adding a classification layer. Voxel wise classification was done on test data volumes by selecting patches centered on every voxel to create a label image. 
The reconstruction error map was obtained for the entire volume using ND. A binary mask derived from the error map, indicating the lesion regions, was used to reject false positives in the label image. 
\subsection {Preprocessing}
\subsubsection{Histogram Matching}
\begin{figure*}

\subfloat[]{\includegraphics[width=0.33\textwidth]{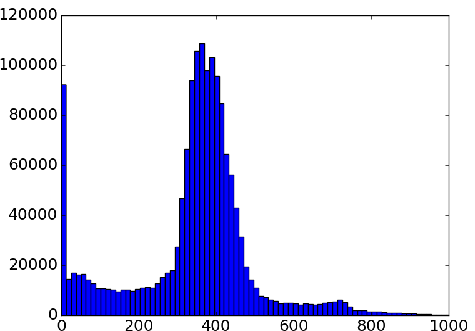}}\hfill
\subfloat[]{\includegraphics[width=0.33\textwidth]{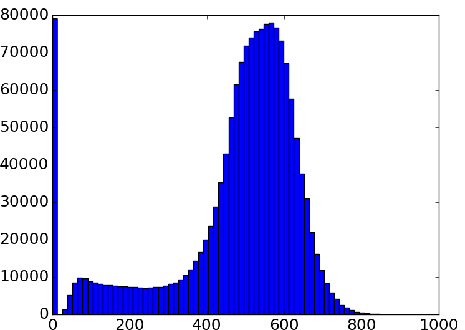}} \hfill
\subfloat[]{\includegraphics[width=0.33\textwidth]{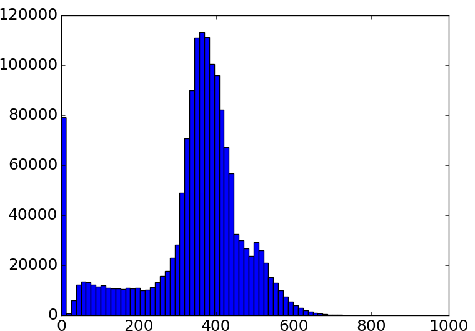}} \\


\caption{Histogram Matching (a) Histogram of reference FLAIR sequence. (b) Histogram of Test FLAIR sequence. (c) Histogram of test data post Histogram Matching.  }
\label{fig:data}
\end{figure*}

All the volumes in the database were histogram matched \cite{IEEEhowto:histmatch} to an arbitrarily chosen reference image from the training data. This  ensures the contrast and dynamic range to be similar across image volumes, (Fig. \ref{fig:data} (a-c)). The same reference image was used for  HGG, LGG and Ischemic data set.
\subsubsection{ z-score}
Following histogram matching, all sequences corresponding to a patient volume were independently normalized to have zero mean and unit standard deviation. 
 
\subsection {Patch Extraction}
Patches of size 21x21 were drawn from all four sequences for pre-training and fine tuning the networks.  
\subsubsection{ Patches for SDAE} For pre-training, patches were sampled using a sliding window of 21x21 with a stride of 10 throughout the image volume, ignoring the voxel labels. Patches for fine tuning  were extracted from regions around the tumor. This sampling scheme reduces the data imbalance between lesion and non-lesion patches. The patch extraction scheme for the network is shown in Table \ref{patch_scheme}.
\par Since the SDAEs were pre-trained using unlabeled data and fine tuned with limited labeled data the networks are referred to as Deep Semi-supervised network (D-SSN).
\subsubsection{Patches for ND} The patches for training the ND were extracted from FLAIR and T2 images. The non-lesion regions were sampled to obtain the patches.  
 
\begin{table}[h]
\centering
\caption{Patch Extraction Scheme for D-SSN}
\label{patch_scheme}
\begin{tabular}{l|l|l|l|}
\cline{2-4}
                           & Pre-training                                                                     & Fine-Tuning                                                                   & No. patients                                                                                                                 \\ \hline
\multicolumn{1}{|l|}{D-SSN} & \begin{tabular}[c]{@{}l@{}}Systematic Sampling,\\ No class balance\end{tabular} & \begin{tabular}[c]{@{}l@{}}Vicinity of Tumor,\\ No class balance\end{tabular} & \begin{tabular}[c]{@{}l@{}}Pre-training=135\\ Fine-tuning=20\\\end{tabular} \\ \hline
\end{tabular}
\end{table}

\subsection {Training}
\subsubsection{Novelty Detector}

The Novelty detector is a one layer deep DAE, (Table \ref{Narchi}), with a sigmoid encoding layer and a linear decoder. The novelty detector was trained on  1,110,492 patches (576636-ST; 533856-LT) and validated on 438,275 patches (193955-ST; 244320-LT) extracted from the same subset of data that was used fine tune the HGG network. The training data was corrupted by 20 \% masking noise. The weights and biases of the network were randomly initialized. The network was trained for 200 epochs with an initial learning rate of 0.001. Mean squared error loss function with L2 regularization was optimized with RmsProp \cite{IEEEhowto:Rmsprop}. 


\subsubsection {Stacked Denoising Autoencoders}
 Separate networks with the same architecture were trained for LGG and HGG segmentation. The network architecture is given in Table \ref{Narchi}. Both the networks were pre-trained using 130 HGG patients (70 ST and 60 LT data). The HGG network was fine tuned using patches from 10 ST images and 10 LT images. Validation was done using patches from 11 ST and 10 LT. The data set for fine tuning was a subset of data set used for pre- training. The LGG network was fine-tuned using 20 patient image volumes and validated using patches from 11 patient volumes.
\begin{table}[h]

\caption{Network Architecture. H$_i$- No. of Neurons in the i$^{th}$ Hidden Layer}
\label{Narchi}
\centering
\begin{adjustbox}{width=0.48\textwidth}
\small
\begin{tabular}{|c|c|c|c|c|c|c|}
\hline
Network & Input Layer & H$_1$ & H$_2$ & H$_3$ & H$_4$ & Output Layer\\
\hline
ND &882&3500&&&&882\\
\hline
SDAE &1764 & 3500 & 2000& 1000 &500 & 5 \\
\hline

\end{tabular}
\end{adjustbox}
\end{table}
\subsubsection{HGG Network}
The HGG network was pre-trained layer by layer using  941,716 patches with 25\% masking noise for 50 epochs. The weights \& biases in each layer was initialized using Xavier initialization \cite{IEEEhowto:xavier} and RmsProp was used as the optimizer. The networks used a sigmoid encoder and a linear decoder.  
 For fine-tuning, the weights and biases connecting the penultimate layer and decision layer was initialized with zeros. The network was trained  using  3,304,035 patches (18,88,020-ST and 14,16,015-LT) and validated on 411,495 patches(235,140-ST; 176,355-LT). 
The weights of the network were learnt by minimizing the  negative log likelihood cost function using Stochastic Gradient Descent with momentum equal to 0.9. The learning rate was initialized to 0.005 and was annealed as a function of number of epochs (Eq. \ref{eq:learning_rate}) with a learning rate decay of 0.001. To prevent overfitting, all layers used dropouts \cite{IEEEhowto:Dropout} of 25 \%.
\begin{eqnarray}
current\,learning\,rate=\frac{initial\,learning\,rate}{epoch*learning\,rate\,decay} 
\label{eq:learning_rate}
\end{eqnarray}

\subsubsection{LGG Network}

 Due to the limited amount of LGG volumes in the data set, the network pre-trained on the HGG data was fine tuned with dropouts (35\%) using LGG image patches(Training- 1,365,450; validation- 181,170). The data augmentation scheme described later increases the training data 6 fold.
 
\subsection {Data augmentation}
Lesion classes constitute less than 2\% of the image volume which makes data augmentation unavoidable. Data augmentation was done by rotating image patches through various angles. The angles were chosen such that the fill in regions are minimized during the interpolation. Arbitrary angles are also possible but the impact of zero filling or zero padding would be difficult to determine. Augmentation is done on the fly, thus minimizing hard disk and RAM usage. Multi-threaded training ensured that patches were augmented and loaded into GPU memory without slowing down training. It was observed that performing label preserving rotations during fine tuning had a significant impact on the classifying less prevalent classes like Non Enhancing Tumor and Necrotic Region.
\par For the HGG network, patches extracted from single time points were rotated by either 90, -90 or 180 degrees, while patches from longitudinal data were rotated by all three angles. For the LGG network, patches were additionally rotated by -45 \& 45 degrees.
\subsection { Hyper-parameter optimization}
Hyper-parameters were set by random search in the space of hyper-parameters. The training patch size was included as a hyper-parameter, in addition to the learning rate, optimizer, number of layer, number of neurons per layer and the L1/L2 penalties. Networks trained with various combinations of hyper-parameters were tested on limited test data and the hyper-parameters corresponding to the best dice scores were identified. Initially networks were trained with 3D patches which seemed a natural choice for the problem. However, in order to keep the size of the network to a manageable level, 2D patches were adopted. 

\subsection {Postprocessing using ND}
In the test phase for both the networks, vectorized patches (21x21x4)  were used as input to classify the centre voxel of the patch.
\par Patches from the T2 and FLAIR was used as input to the ND. The  reconstruction error map for a slice, $ND$, was constructed by assigning to every voxel( $i_c, j_c$), the mean reconstruction error of the patch centered at that voxel, (Eq. (\ref{eq:nd})). This led to a heat map like image with large error regions corresponding to the location of the Glioma/lesion. In Eq. (\ref{eq:nd}) $p$ is the size of the patch i.e. each patch was of size $p \times p$, $N=  2\times p \times p$  and  $E$ is the patch error. Patch Error, (Eq. (\ref{eq:pe})), is the squared error between the FLAIR ($F$) and $T_2$ ($T$) patches and their respective reconstruction $RF$ and $RT$. The error map was then binarized using Otsu's thresholding \cite{IEEEhowto:Otsu} technique. 
\begin{equation}
ND(i_c,j_c)= \frac{1}{N}\hspace{-0.25 cm}\sum_{i=i_c -(p-1)/2;}^{i_c +(p-1)/2;} \sum _{j=j_c -(p-1)/2;}^{j_c +(p-1)/2} \hspace{-0.15 cm}{E(i-xo,j-yo)}
\label{eq:nd}
\end{equation}
where $xo= i_c-(p-1)/2$ and $yo= j_c-(p-1)/2$
\begin{equation}
E=(RF-F)^2+(RT-T)^2
\label{eq:pe}
\end{equation}

\par Following the generation of binary mask, connected component analysis was carried on the image predicted by the HGG and LGG networks. Connected components that had a non-empty intersection with the binary error mask were retained while the rest were discarded. 

\par A variant of the aforementioned Novelty Detector  called Cascaded Novelty Detector (CND) was developed. In CND, the final resultant error value corresponding to a voxel was calculated by maintaining a cumulative error sum over all the image patches containing the voxel. The calculation of the error for each voxel is given in Eq. (\ref{eq:cascaded}), where $i$ and $j$ are the voxel coordinates of the error map in a given slice, p is the size of the patch \& $E$ is the patch error.
\begin{equation}
CND(i,j)=\sum_{i_c= i-(p-1)/2;}^{i+(p-1)/2;}\sum_{ j_c=j-(p-1)/2;}^{j+(p-1)/2;}\hspace{-0.8cm}{E(i-xo,j-yo)}
\label{eq:cascaded}
\end{equation}

\subsection{Submission to BraTS 2015 challenge} The authors submission to the BraTS 2015 challenge is described in \cite{IEEEhowto:sdae1}. Briefly, two SDAEs with 3 hidden layers (3000-1000-500) each, were trained on 3D image patches. One using HGG data and one using a mix of HGG and LGG data. Pre-processing included histogram matching, z score normalization and intensity clipping. During pre-training, class labels were used to form balanced mini-batches. The results of the prediction from these two networks were combined to obtain the label image. The label image was registered to an anatomic atlas to remove connected connected components in anatomical regions where the probability of tumor occurrence is generally considered low. The largest connected component was retained.



\begin{figure*}

\subfloat[]{\includegraphics[width=0.24\textwidth,keepaspectratio]{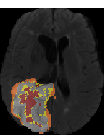}}\hfill
\subfloat[]{\includegraphics[width=0.24\textwidth,keepaspectratio]{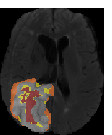}} \hfill
\subfloat[]{\includegraphics[width=0.24\textwidth,keepaspectratio]{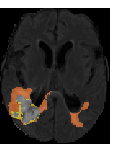}}\hfill
\subfloat[]{\includegraphics[width=0.24\textwidth,keepaspectratio]{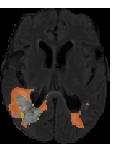}} \\
\subfloat[]{\includegraphics[width=0.24\textwidth,keepaspectratio]{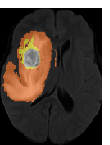}}\hfill
\subfloat[]{\includegraphics[width=0.24\textwidth,keepaspectratio]{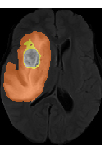}} \hfill
\subfloat[]{\includegraphics[width=0.24\textwidth,keepaspectratio]{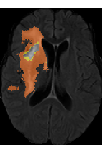}}\hfill
\subfloat[]{\includegraphics[width=0.24\textwidth,keepaspectratio]{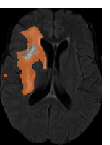}} \\

\subfloat[]{\includegraphics[width=0.24\textwidth,keepaspectratio]{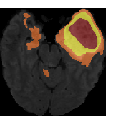}} \hfill
\subfloat[]{\includegraphics[width=0.24\textwidth,keepaspectratio]{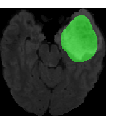}} \hfill
\subfloat[]{\includegraphics[width=0.24\textwidth,keepaspectratio]{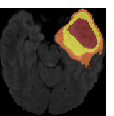}} \hfill
\subfloat[]{\includegraphics[width=0.24\textwidth,keepaspectratio]{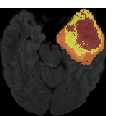}}\\

\subfloat[]{\includegraphics[width=0.24\textwidth,keepaspectratio]{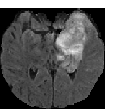}} \hfill
\subfloat[]{\includegraphics[width=0.24\textwidth,keepaspectratio]{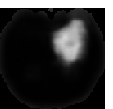}} \hfill
\subfloat[]{\includegraphics[width=0.24\textwidth,keepaspectratio]{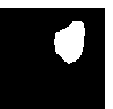}} \hfill
\subfloat[]{\includegraphics[width=0.24\textwidth,keepaspectratio]{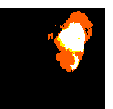}}\\




\caption{Performance of Proposed Networks. (a) Ground Truth. (b) Prediction. (c) Ground Truth. (d) Prediction. (e) Ground Truth. (f) Prediction. (g) Ground Truth. (h) Prediction. (i) Raw Prediction. (j)  Otsu's Mask. (k) Prediction after Post Processing. (l) Ground Truth. (m) FLAIR. (n) ND Error Map. (o) Binarized Error Map. (p) Ground Truth. In images all images,  Orange-Edema, Yellow -Non Enhancing Tumor, Red-Necrotic Region, White-Enhancing Tumor, Green-Otsu's Mask. In image (o), White- Binarized Error Map.}
\label{fig:compa}
\end{figure*}

\begin{figure*}
\subfloat[]{\includegraphics[width=0.20\textwidth,keepaspectratio]{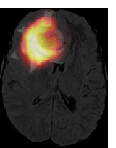}}\hfill
\subfloat[]{\includegraphics[width=0.20\textwidth,keepaspectratio]{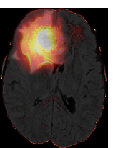}} \hfill
\subfloat[]{\includegraphics[width=0.20\textwidth,keepaspectratio]{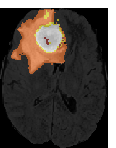}}\hfill
\subfloat[]{\includegraphics[width=0.20\textwidth,keepaspectratio]{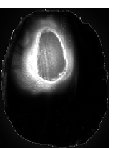}}\hfill
\subfloat[]{\includegraphics[width=0.20\textwidth,keepaspectratio]{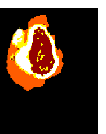}} \\

\subfloat[]{\includegraphics[width=0.20\textwidth,keepaspectratio]{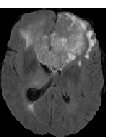}}\hfill
\subfloat[]{\includegraphics[width=0.20\textwidth,keepaspectratio]{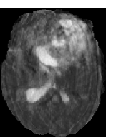}} \hfill
\subfloat[]{\includegraphics[width=0.20\textwidth,keepaspectratio]{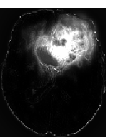}}\hfill
\subfloat[]{\includegraphics[width=0.20\textwidth,keepaspectratio]{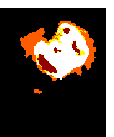}}\hfill
\subfloat[]{\includegraphics[width=0.20\textwidth,keepaspectratio]{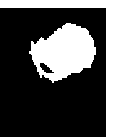}} \\

\subfloat[]{\includegraphics[width=0.20\textwidth,keepaspectratio]{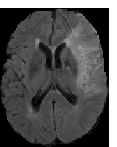}}\hfill
\subfloat[]{\includegraphics[width=0.20\textwidth,keepaspectratio]{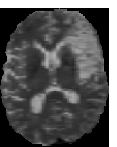}} \hfill
\subfloat[]{\includegraphics[width=0.20\textwidth,keepaspectratio]{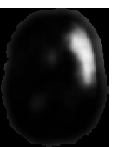}}\hfill
\subfloat[]{\includegraphics[width=0.20\textwidth,keepaspectratio]{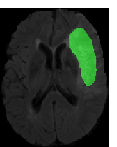}}\hfill
\subfloat[]{\includegraphics[width=0.20\textwidth,keepaspectratio]{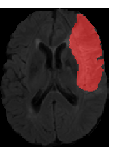}} \\

\subfloat[]{\includegraphics[width=0.20\textwidth,keepaspectratio]{Krish20.eps}}\hfill
\subfloat[]{\includegraphics[width=0.20\textwidth,keepaspectratio]{Krish21.eps}} \hfill
\subfloat[]{\includegraphics[width=0.20\textwidth,keepaspectratio]{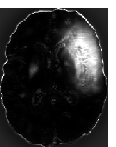}}\hfill
\subfloat[]{\includegraphics[width=0.20\textwidth,keepaspectratio]{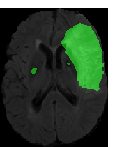}}\hfill
\subfloat[]{\includegraphics[width=0.20\textwidth,keepaspectratio]{Krish24.eps}} \\



\caption{ Performance of ND \& CND on BraTS (a-j) and ISLES data set (k-t). (a) ND Error Map. (b) CND Error Map. (c) Ground Truth. (d) CND Error Map. (e) Ground Truth. (f) FLAIR. (g) T2. (h) CND Error Map. (i) Ground Truth. (j) Binarized CND error map. (k) FLAIR. (l) T2. (m) ND Error Map. (n) Binarized ND error map. (o) Ground Truth. (p) FLAIR. (q) T2. (r) CND error map. (s) Binarized Error Map. (t) Ground Truth. In  images (c), (e) and (i) , Orange- Edema, Yellow- Non Enhancing Tumor, Red- Necrotic Region,White- Enhancing Tumor. In images (n), (o), (s) and (t), Green-Binarized Error Map, Red-Ischemic Lesion Ground truth}
\label{fig:nd_out}
\end{figure*}

\section{Results and Discussion}

\subsection {Performance on the Training data}
The dice scores on the single time point patients, longitudinal patients and the entire BraTS 2015 training data set is shown in Table \ref{dssn STATS}. The performance of algorithm on the single time point patients is shown in Fig. \ref{fig:compa} (a-d).
\par Fig. \ref{fig:compa} (e-h) shows the performance of the network on a longitudinal patient data at 2 different time points.
By including longitudinal time point patients as training data, the network attained the capability to capture the tumor region across time points. 
\begin{table}[h!]
\caption{D-SSN Performance on HGG data. LT and ST refer to longitudinal and single time point image volumes. WT- Whole Tumor Dice score, TC- Tumor Core Dice Score, AT- Active Tumor Dice Score, $\mu$- Mean, $\sigma$- Standard Deviation, $M$- Median, $n$- No. of patient volumes. }
\centering
\begin{adjustbox}{width=0.48\textwidth}
\small
\begin{tabular}{|c|c|c|c|c|}
\hline
Grade &Statistics &WT  & TC & AT \\
\hline
\multirow{3}{*}{All$(n= 220)$ } &M  &  0.89 & 0.81 & 0.84   \\

& $\mu$   & 0.85    &0.71   & 0.75 \\

&$\sigma$      & 0.11  & 0.25  &0.22 \\
\cline{2-5}

\multirow{3}{*}{ST  $(n=123)$} &M  &  0.92 & 0.86 & 0.87   \\
&$\mu$   & 0.86    &0.78   & 0.80 \\
&$\sigma$      & 0.13  & 0.20  &0.18 \\
\cline{2-5}

\multirow{3}{*}{LT $(n = 97)$} &M  &  0.85 & 0.70 & 0.79   \\
&$\mu$   & 0.84    &0.62   & 0.69 \\
&$\sigma$      & 0.08  & 0.27  &0.25 \\
\cline{2-5}
\hline

\end{tabular}
\label{dssn STATS}
\end {adjustbox}
\end{table}




\subsection {Novelty detector}
Fig. \ref{fig:compa} (i-l) demonstrates the reduction in false positive voxels using ND. Post processing using the ND mask led to good improvements in glioma segmentation. The improvement in performance was in the order of 4\% for HGG whole tumor dice score, 1 \% for HGG tumor core and 1\% for HGG active tumor and 3 \% for LGG whole tumor dice score. 
\par Fig. \ref{fig:compa} (n) shows the reconstruction error heat map for a sample slice, Fig. \ref{fig:compa} (m). A bi-modal distribution could be inferred on visual inspection of the error map. The mean square reconstruction error corresponds to maximizing the log-likelihood of the training data assuming a Gaussian distribution. Given this interpretation any data input that does not correspond to the training data distribution can be expected to give rise to a large mean square error enabling lesion patch detection. Cross-validation can be used to determine the ideal threshold but would have to be changed depending on the lesion. Otsu's thresholding technique makes the ND application independent.
\par A sample ND error map binarized using Otsu's thresholding is shown in Fig. \ref{fig:compa} (o). The mean square error of a patch is assigned to the center pixel of the patch, consequently the voxels towards the boundaries of the lesions will get assigned a much lower error than the voxels near the center of the lesion. Thus the ND error map underestimate the size of the lesion. For every voxel in the volume, CND's take into account the reconstruction error from patches centered on its neighbors, therefore the degree of under-segmentation of the lesion is lower in CND's when compared to ND's (Fig. \ref{fig:nd_out} (a-c)). 
\par Qualitatively, it was observed that CND's produces unique error distribution for various constituents of the lesion (Fig.\ref{fig:nd_out} (d-e), (h-i)). From the CND error map, the necrotic region would be easily delineated from edema and enhancing tumor. It is note worthy that even though CND was trained on FLAIR and T2 and not T1c, it was able to delineate enhancing tumor regions. However, this was only possible if there existed a corresponding hyper intensity profile in either of its input sequences. Thus CND trained with FLAIR, T2 and T1 post contrast would be the ideal choice to capture all the constituents of the lesion. 
\par The CND generated error map could be used as initialization point for various generative techniques. The whole lesion could be segmented from the CND error map by setting the threshold to be one standard deviation away from the mean (Fig. \ref{fig:nd_out} (j)).
 
\par The performance of ND on ISLES challenge data is shown in Fig. \ref{fig:nd_out} (k-o). Similar to Glioma segmentation, the error map picks up the location of the ischemic lesions accurately, however ND misses lesions that constitute less than 1\% of total number of voxels in the volume. The T2 weighted sequences in the ISLES data set were acquired in the sagittal plane as opposed to the axial plane acquisition in the BraTS data set. Since the ND was trained using BraTS data set, the resolution mismatch would lead to poor performance in detecting small lesions. On patients with lesions that constitute more than 1\% of the total number of voxels, ND achieved a dice score of 0.44 $\pm$ 0.21. In contrast, CND achieved a much higher dice score of 0.64 $\pm$ 0.17, (Fig. \ref{fig:nd_out} (p-t)). The improved performance was due to the reasons explained in the previous paragraph(s). These results imply that the ND can be trained using data from healthy volunteers or other imaging studies comprising of relevant MR sequences.

\subsection{Transfer Learning for LGG}
The LGG network was trained by fine tuning the pre-trained HGG network. There was a significant improvement in network performance compared to the authors submission to the BraTS 2015 challenge. Comparison with networks pre-trained with a mix of LGG and HGG data are shown in Table \ref{dssnvsprevious}. However, the performance of the LGG network is significantly worse than of the HGG network due to inherent differences in the abundance of classes found in these grades of Glioma. Enhancing tumor or active tumor region is hardly present in LGG while non-enhancing tumor class is rare in HGG. A collection of unlabeled multi-sequence LGG volumes can be expected to improve the prediction accuracy significantly. 



\subsection{Prediction with missing sequences} 
The performance of the network upon blocking individual sequences from the input is shown in Table \ref{Missing Sequence}. Prediction with a missing sequence was expected to lower the dice scores, however the magnitude of the decline was dependent on the sequence dropped. The results were also informative, indicating the relative importance of the sequences. Removing T1 had negligible impact on the whole tumor score while removing T2 and FLAIR lead to the maximum change i.e. decline. The change in dice scores of enhancing tumor or active tumor was the largest when T1c was removed which can be expected. Based on the decline in performance  one can conclude that FLAIR  plays an important role in delineating lesion from normal tissues.



\begin{table}
\caption{Performance of D-SSN Performance with Missing Sequences (MS), FLAIR(FL), T2, T1, T1 post contrast(T1c). WT- Whole tumor Dice score, TC- Tumor Core Dice Score, AT- Active Tumor Dice Score, $\mu$- Mean, $\sigma$- Standard Deviation, $M$- Median.}
\centering
\begin{adjustbox}{width=0.49\textwidth}
  \begin{tabular}{|c|c|c|c|c|c|c|c|c|c|}
  \hline
    \multirow{1}{*}{MS} &
      \multicolumn{3}{c|}{WT} &
      \multicolumn{3}{c|}{TC} &
      \multicolumn{3}{c|}{AT} \\
      
    & $\mu$ & $\sigma$ & $M$ & $\mu$& $\sigma$ & $M$ & $\mu$ & $\sigma$ & $M$\\
    \hline
    FL & 0.35& 0.23 & 0.35 & 0.55 &0.29 &0.63 & 0.58 &0.31 &0.68\\
    \hline
    T2 & 0.79& 0.13 & 0.82 & 0.61 &0.29 &0.69 & 0.70 &0.26 &0.80\\
    \hline
    T1 & 0.80& 0.16 & 0.85 & 0.61 &0.25 &0.66 & 0.58 &0.29 &0.60\\
    \hline
    T1c & 0.81& 0.13 & 0.86 & 0.40 &0.22 &0.38 & 0.00 &0.00 &0.00\\
    \hline
  \end{tabular}
  \label{Missing Sequence}
  \end{adjustbox}
\end{table}
\subsection {Weakly supervised learning}
The minimum amount of data required for the network to maintain its level of performance was tested by fine tuning the LGG and HGG networks with a lower number of patient data (leading to decreasing number of patches). The results in Table \ref{data_size} shows that training with patches drawn from only 20 patients, the  HGG network had marginal decline in dice scores and are comparable to results obtained when the networks were trained on patches drawn from a larger number of patients. It's also notable that if the number of extracted patches are increased from a limited number of patients then the network performance rebounds as shown in Table \ref{data_size}. The structures in the brain appear similar across different brain MR images. Drawing patches from a limited number of patient volumes coupled with data augmentation, would still provide enough samples for the network to learn and maintain prediction performance.

\begin{table}
\caption{Performance of network based on number of Training patients used (N), (20M) is 20 patients with more patches. WT- Whole tumor Dice score, TC- Tumor Core Dice Score, AT- Active Tumor Dice Score, $\mu$- Mean, $\sigma$- Standard Deviation, $M$- Median. }
\centering
\begin{adjustbox}{width=0.49\textwidth}
  \begin{tabular}{|c|c|c|c|c|c|c|c|c|c|}
  \hline 
    \multirow{1}{*}{N} &
      \multicolumn{3}{c|}{WT} &
      \multicolumn{3}{c|}{TC} &
      \multicolumn{3}{c|}{AT} \\
    & $\mu$ & $\sigma$ & $M$ & $\mu$ & $\sigma$ & $M$ & $\mu$ & $\sigma$ & $M$\\ 
    \hline 
    20 & 0.84& 0.13 & 0.89 & 0.72 &0.24 &0.81 & 0.74 &0.25 &0.84\\ 
    \hline
    40 & 0.85 & 0.13 & 0.90 & 0.75 &0.23 &0.83 & 0.78 &0.23 & 0.87\\
    \hline
    65 & 0.84& 0.15 & 0.89 & 0.75 &0.23 &0.83 & 0.78 &0.24 &0.86\\
    \hline
    20M & 0.86 & 0.12 & 0.90 & 0.73 &0.24 &0.83 & 0.77 &0.23 &0.86\\
    \hline
  \end{tabular}
  \end{adjustbox}
  \label{data_size}
\end{table}
\subsection{Performance on Challenge Dataset}

The networks were tested on two different challenge test data namely BraTS 2013 challenge test data and BraTS 2015 test data. The performance of networks on BraTS 2013 challenge data and BraTS 2015 test data is given in Table \ref{dssnvsprevious}. On the BraTS 2013 leader-board, the network was ranked $8^{th}$.

\par Compared to the authors previous submission to the 2015 challenge, on HGG data it was observed that the current method does significantly better on the tumor core as well as the active tumor while the same level of performance was maintained for the whole tumor dice score. For LGG, the use of a mix of LGG and HGG patches to pre-train and fine tune the network gave better results for whole tumor. However the transfer learning approach gives significantly better results for the tumor core. These results  indicate that an  optimum mix of LGG and HGG data is required for improved segmentation performance on LGG patient volumes.

\begin {table} [h!]
\caption {Performance of D-SSN on challenge data sets compared against the authors original submission to Brats 2015 challenge. Nw- Network, G- Grade of Tumor, WT- Whole Tumor Dice score, TC- Tumor Core Dice score, AT- Active Tumor Dice score. PS- Previous Submission, D-SSN-Current Submission.}
\label {dssnvsprevious}
\centering
\begin{adjustbox}{width=0.499\textwidth}
\small
\begin {tabular}{|c|c|c|c|c|c|}
\hline
Year&Nw&G&WT &TC & AT \\
\hline
2013 & D-SSN & All &0.85 $\pm$0.04&0.78$\pm$0.15 & 0.73$\pm$0.11\\
\hline
\multirow{6}{*}{2015} &\multirow{3}{*}{ PS} &All &0.71$\pm$0.24&0.51$\pm$0.26&0.58$\pm$0.17 \\
\cline{3-6}
&  & HGG &0.71$\pm$0.23&0.57$\pm$0.24&0.58$\pm$0.17\\
\cline{3-6}
 & & LGG & 0.73$\pm$0.29 & 0.38$\pm$0.28 & *\\
\cline{2-6}
&\multirow{3}{*}{ D-SSN} & All & 0.73$\pm$0.25 &0.56$\pm$0.28&0.68$\pm$0.20\\
\cline{3-6}
& & HGG & 0.75$\pm$0.19 &0.61$\pm$0.27&0.68$\pm$0.20\\
\cline{3-6}
&  & LGG & 0.68$\pm$0.34 &0.46$\pm$0.30&*\\
\hline

\end {tabular}
\label {challenge}
\end{adjustbox}
\end {table}

\section{Conclusion}

In this paper we propose a completely automated brain tumor segmentation technique with a novel false positive/candidate detection method based on denoising autoencoders.

\begin {itemize}
\item Despite differences in acquisition resolution, ND trained using non-lesion patches (BraTS data) was able to learn the normal brain structure and detect ischemic lesions (ISLES data). A variant of ND (CND), wherein a cumulative error map was calculated for every voxel, was able to significantly improve lesion detection performance on ISLES data. In addition, CND error maps assigned different error distributions to various constituents of glioma, making it an ideal tool to construct tumor atlases. This can also serve as a good initialization for various segmentation techniques. 

\item The paper clearly demonstrates the ability of SDAE's to produce good segmentation using minimal number of patient data. The redundancy of patches obtained from MR brain images was exploited to train networks.
\item  The results presented are the prediction of a single network with minimal data pre-processing and post-processing. The N4 bias correction technique which is an oft used pre-processing step was eliminated. Histogram matching to a reference data was still done and future work would be to eliminate the same by appropriate data normalization. Skull stripping (BraTS data and ISLES challenge data were skull stripped) could potentially be eliminated as a separate step by using ND. The idea is to enable prediction on MR images without expensive pre-processing.

\end {itemize}

\par  In summary the work presented applies SDAE's for the brain lesion detection and segmentation task using a limited number of training data. The novelty detector concept allows for efficient elimination of false positives and candidate detection, making it a valuable CAD tool.


\ifCLASSOPTIONcaptionsoff
  \newpage
\fi

\end{document}